\documentclass[11pt,a4paper]{article}
\usepackage[hyperref]{AACL-IJCNLP2020}
\usepackage{times}
\usepackage{latexsym}
\usepackage{array}

\usepackage{graphicx}
\usepackage{amsmath,amsfonts,amssymb}
\usepackage{mathtools}
\usepackage{caption}
\usepackage{lipsum}
\usepackage{booktabs}
\usepackage{xcolor}
\usepackage{color, colortbl}
\definecolor{Gray}{gray}{0.9}
\definecolor{ao(english)}{rgb}{0.0, 0.5, 0.0}
\definecolor{cardinal}{rgb}{0.77, 0.12, 0.23}
\usepackage{url}
\usepackage{float}
\usepackage{tabularx}
\usepackage{appendix}
\usepackage{placeins}
\makeatletter
\newcommand{\ssymbol}[1]{^{\@fnsymbol{#1}}}
\makeatother

\usepackage{microtype}

\aclfinalcopy 

\title{Improving Context Modeling in Neural Topic Segmentation}

\author{First Author \\
  Affiliation / Address line 1 \\
  Affiliation / Address line 2 \\
  Affiliation / Address line 3 \\
  \texttt{email@domain} \\\And
  Second Author \\
  Affiliation / Address line 1 \\
  Affiliation / Address line 2 \\
  Affiliation / Address line 3 \\
  \texttt{email@domain} \\}

\author{Linzi Xing$^\dagger$, Brad Hackinen$^\ddagger$, Giuseppe Carenini$^\dagger$, Francesco Trebbi$^\mathsection$\\
  $\dagger$ University of British Columbia, Vancouver, Canada \\
  $\ddagger$ Ivey Business School, London, Canada \\
  $\mathsection$ University of California Berkeley, California, USA \\
  \tt \{lzxing, carenini\}@cs.ubc.ca \\
  \tt bhackinen@ivey.ca \\ 
  \tt ftrebbi@berkeley.edu
  }

\date{}

\begin{document}
\maketitle
\begin{abstract}

    Topic segmentation is critical 
    in key NLP tasks and recent works favor highly effective neural supervised  approaches.
    However, current neural solutions are arguably limited in how they model context.
    In this paper, we enhance a segmenter based on a hierarchical attention BiLSTM network to better model context, by adding a coherence-related auxiliary task and restricted self-attention. Our optimized segmenter\footnote{Our code will be publicly available at \url{www.cs.ubc.ca/cs-research/lci/research-groups/natural-language-processing/}} outperforms SOTA approaches when trained and tested on three datasets. We also the robustness of our proposed model in domain transfer setting by training a model on a large-scale dataset and testing it on four challenging real-world benchmarks. Furthermore, we apply our proposed strategy to two other languages (German and Chinese), and show its effectiveness in multilingual scenarios.
\end{abstract}

\section{Introduction}

Topic segmentation is a fundamental NLP task that has received considerable attention in recent years \cite{barrow-etal-2020-joint, glavas-2020-two, cross_attn_2020}. 
It can reveal important aspects of a document semantic structure by splitting the document into topical-coherent textual units.
Taking the \textit{Wikipedia} article in Table~\ref{tab:example} as an example, without the section marks, a reliable topic segmenter should be able to detect the correct boundaries within the text and chunk this article into the topical-coherent units \texttt{T1}, \texttt{T2} and \texttt{T3}.
The results of topic segmentation can further benefit other key downstream NLP tasks such as document summarization \cite{mitra-etal-1997-automatic, riedl-biemann-2012-text, wen-2019-extractive}, question answering \cite{oh07, dennis-2017-core}, machine reading \cite{1981, saha-2019-aligning} and dialogue modeling \cite{topic_dialogue_2020, ijcai2020-517}.

\begin{table}
\centering
\scalebox{0.8}{
 \begin{tabular}{|m{23em}|} 
 \hline
 \rowcolor{Gray}
 \underline{\textbf{\textit{Preface:}}} \\
 \rowcolor{Gray}
 Marcus is a city in Cherokee County, Iowa, United States. \\
 \hline\hline
 \underline{\textbf{[T1] \textit{History:}}} \\
 \underline{S1}: The first building in Marcus was erected in 1871.\\
 \underline{S2}: Marcus was incorporated on May 15, 1882. \\
 \hline
 \underline{\textbf{[T2] \textit{Geography:}}} \\
 \underline{S3}: Marcus is located at (42.822892, -95.804894).\\
 \underline{S4}: According to the United States Census Bureau, the city has a total area of 1.54 square miles, all land. \\
 \hline
 \underline{\textbf{[T3] \textit{Demographics:}}} \\
 \underline{S5}: As of the census of 2010, there were 1,117 people, 494 households, and 310 families residing in the city. \\
 ... ...\\
 \hline
\end{tabular}
}
\caption{\label{tab:example} 
A Wikipedia sample article about \textit{City Marcus} covering three topics: \texttt{T1}, \texttt{T2} and \texttt{T3}}
\end{table}

A wide variety of techniques have been proposed for topic segmentation.
Early unsupervised models exploit word statistic overlaps \cite{hearst-1997-text, Galley:2003}, Bayesian contexts \cite{eisenstein-barzilay-2008-bayesian} or  
semantic relatedness graphs \cite{glavas-etal-2016-unsupervised} to measure the lexical or semantic cohesion between the sentences or paragraphs and infer the segment boundaries from them.
More recently, several works have framed topic segmentation as neural supervised learning, because of the remarkable success achieved by such models in most NLP tasks \cite{wang-etal-2016-learning, wang-etal-2017-learning, Imran-2017, koshorek-etal-2018-text, arnold2019sector}. 
Despite 
minor architectural differences, most of these neural solutions adopt Recurrent Neural Network \cite{rnn97} and its variants (RNNs) as their main framework. 
On the one hand, RNNs are appropriate because topic segmentation can be modelled as a sequence labeling task where each sentence is either the end of a segment or not. On the other hand, this choice makes these neural models limited in how to model the context. Because some sophisticated RNNs (eg., LSTM, GRU) are able to preserve long-distance information \cite{rnnreview14, Imran-2017, wang-etal-2018-toward}, which can largely help language models. But for topic segmentation, it is critical to supervise the model to focus more on the local context.  


As illustrated in Table~\ref{tab:example}, the prediction of the segment boundary between \texttt{T1} and \texttt{T2} hardly depends on the content in \texttt{T3}.
Bringing in excessive long-distance signals may cause unnecessary noise and 
hurt 
performance.
Moreover, text coherence has strong relation with topic segmentation \cite{wang-etal-2017-learning, glavas-2020-two}. 
For instance, in Table~\ref{tab:example}, sentence pairs from the same segment (like $<$\texttt{S1}, \texttt{S2}$>$ or $<$\texttt{S3}, \texttt{S4}$>$) 
are more coherent 
than sentence pairs across segments (like \texttt{S2} and \texttt{S3}). 
Arguably, with a proper way of modeling the coherence between adjacent sentences, a topic segmenter can be further enhanced.  


In this paper, we propose to enhance a state-of-the-art (SOTA) topic segmenter \cite{koshorek-etal-2018-text} based on hierarchical attention BiLSTM network to better model the local context of a sentence in two complementary ways.
First, we add a coherence-related auxiliary task to make our model learn more informative hidden states for all the sentences in a document. 
More specifically, we refine the objective of our model to encourage smaller coherence for the sentences from different segments and larger coherence for the sentences from the same segment. 
Secondly, we enhance context modeling by utilizing restricted self-attention \cite{wang-etal-2018-toward}, which enables our model to pay attention to the local context and make better use of the information from the closer neighbors of each sentence (i.e., with respect to a window of explicitly fixed size $k$). 
Our empirical results show (1) that our proposed context modeling strategy significantly improves the performance of the SOTA neural segmenter on three datasets,
(2) that the enhanced segmenter is more robust in domain transfer setting when applied to four challenging real-world test sets, sampled differently from the training data,
(3) that our context modeling strategy is also effective for the segmenters trained on other challenging languages (eg., German and Chinese), rather than just English. 

\section{Related Work}
\label{sec:related-work}

\paragraph{Topic Segmentation} 
Early unsupervised models exploit the lexical overlaps of sentences to measure the lexical cohesion between sentences or paragraphs \cite{hearst-1997-text, Galley:2003, eisenstein-barzilay-2008-bayesian, riedl-biemann-2012-topictiling}. Then, by moving two sliding windows over the text, the cohesion between successive text units could be measured and a cohesion drop would signal a segment boundary. Even if these models do not require any training data, they only show limited performance in practice and are not general enough to handle the temporal change of the languages \cite{huang-paul-2019-neural-temporality}.

More recently, neural-based supervised methods have been devised for topic segmentation because of their more accurate predictions and greater efficiency. One line of research frames topic segmentation as a sequence labeling problem and builds neural models to predict segment boundaries directly.
\citet{wang-etal-2016-learning} proposed a simple BiLSTM model to label if a sentence is a segment boundary or not. They demonstrated that along with engineered features based on cue phrases (eg., `first of all', `second'), their model can achieve marginally better performance than early unsupervised methods. 
Later, \citet{koshorek-etal-2018-text} proposed a hierarchical neural sequence labeling model for topic segmentation and showed its superiority compared with their selected supervised and unsupervised baselines. 
Around the same time, \citet{Badjatiya-2018} proposed an attention-based BiLSTM model to classify whether a sentence was a segment boundary or not, by considering the context around it.
The work we present in this paper can be seen as pushing this line of research even further by encouraging the model to more explicitly consider contextual coherence, 
as well as to prefer more information from the neighbor context through restricted self-attention.

Another rather different line of works first trains neural models for other tasks, and then uses these models' outputs to predict boundaries. \citet{wang-etal-2017-learning} trained a Convolutional Neural Network (CNN) network to predict the coherence scores for text pairs. Sentences in a pair with large cohesion are supposed to belong to the same segment. However, their ``learning to rank" framework asks for the pre-defined number of segments, which limits their model's applicability in practice. 
Our selected framework overcomes this constraint by tuning a confidence threshold during 
the training stage. A sentence with the output probability above this threshold will be predicted as the end of a segment. Following a very different approach, \citet{arnold2019sector} introduced a topic embedding layer into a BiLSTM model. After training their model to predict the sentence topics, the learned topic embeddings can be utilized for topic segmentation. However, one critical flaw of their method is that it requires a complicated pre-processing pipeline, which includes topic extraction and synset clustering, whose errors can propagate to the main topic segmentation task. In contrast, our proposal only requires the plain content of the training data without any complex pre-processing.
\paragraph{Coherence Modeling}
Early works on coherence modeling merely predict the coherence score for documents by tracking the patterns of entities' grammatical role transition \cite{barzilay-lapata-2005-modeling,barzilay-lapata-2008-modeling}. More recently, researchers started modeling the coherence for sentence pairs by their semantic similarities and used them for higher level coherence prediction or even other tasks, including topic segmentation. 
\citet{wang-etal-2017-learning} demonstrated the strong relation between text-pair coherence modeling and topic segmentation. They assumed that (1) a pair of texts from the same document should be ranked more coherent than a pair of texts from different documents; (2) a pair of texts from the same segment should be ranked more coherent than a pair of texts from different segments of a document. With these assumptions, they created a ``quasi" training corpus for text-pair coherence prediction by assigning different coherence scores to the texts from the same segment, different segments but the same document, and different documents.
Then they proposed the corresponding model, and further use this model to directly conduct topic segmentation. Following their second assumption, we propose a neural solution in which by injecting a coherence-related auxiliary task, topic segmentation and sentence level coherence modeling can mutually benefit each other.


\section{Neural Topic Segmentation Model}
Since RNN-based topic segmenters have shown success with high-quality training data, we adopt a state-of-the-art RNN-based topic segmenter enhanced with attention and BERT embeddings as our basic model. Then, we extend such model to make better use of the local context, something that cannot be done effectively within the RNN framework \cite{wang-etal-2018-toward}. In particular, we add a coherence-related auxiliary task and a restricted self-attention mechanisms to the basic model, so that predictions are more strongly influenced by the coherence between the nearby sentences. As a preview of this section, we first define the problem of topic segmentation and introduce the basic model. In the next section, we motivate and describe our proposed extensions. 



\label{sec:solution}
\subsection{Problem Definition}
\label{sec:solution: problem}
Topic segmentation is usually framed as a sequence labeling task. More precisely, given a document represented as a sequence of sentences, our model will predict the binary label for each sentence to indicate if the sentence is the end of a topical coherent segment or not. Formally,

\noindent
\textbf{\underline{Given}}: A document $d$ in the form of a sequence of sentences $\{ s_1, s_2, s_3, ... , s_k \}$.

\noindent
\textbf{\underline{Predict}}: A sequence of labels assigned to a sequence of sentences $\{ l_1, l_2, l_3, ... , l_{k-1} \}$, where $l$ is a binary label, $1$ means the corresponding sentence is the end of a segment, $0$ means the corresponding sentence is not the end of a segment. We do not predict the label for the last sentence $s_k$, since it is always the end of the last segment.

\begin{figure}
\centering
\includegraphics[width=2.95in]{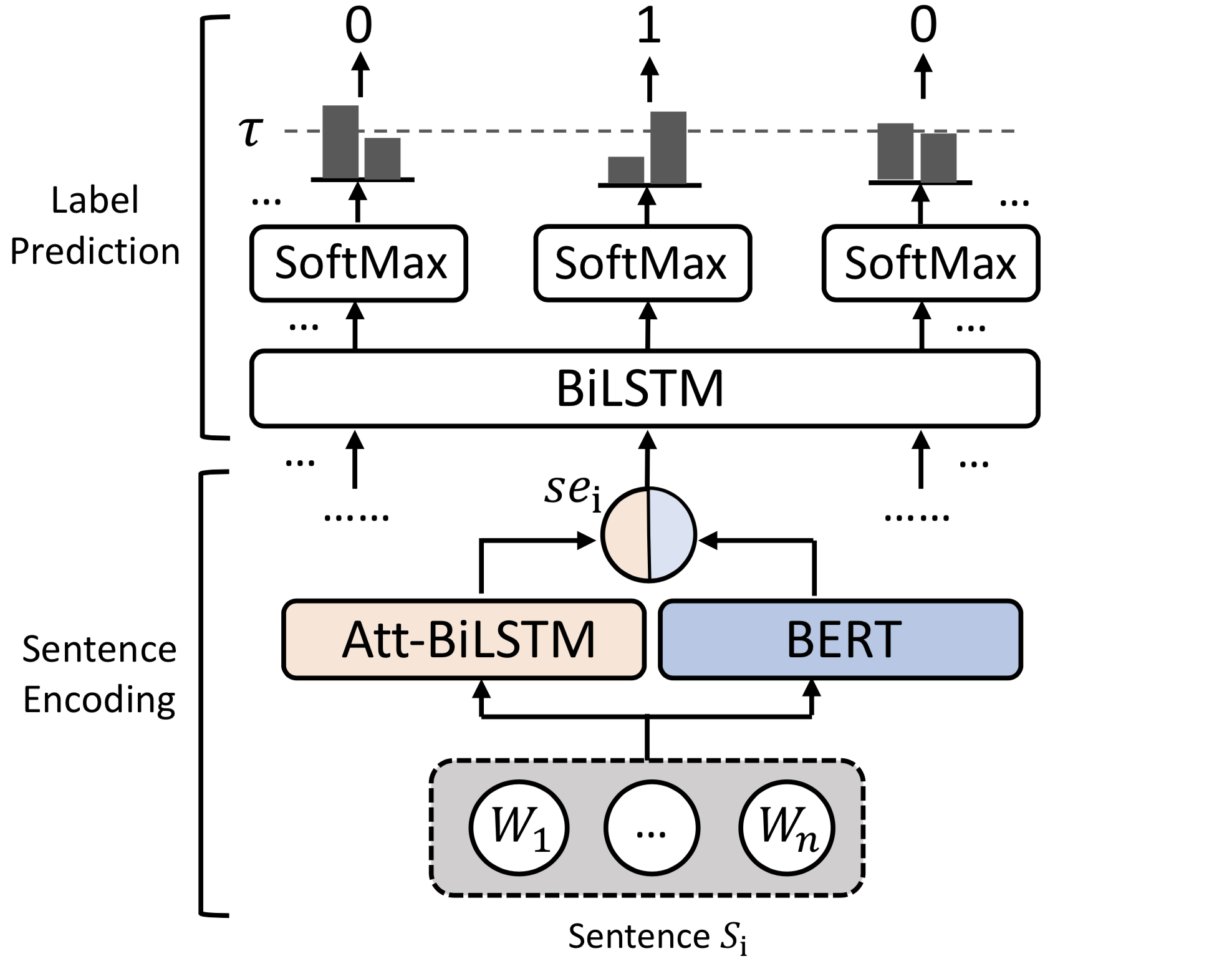}
\caption{\label{fig:basic_model} The architecture of our basic model. $se_{i}$ is the produced sentence embedding for sentence $S_{i}$.}
\end{figure}

\subsection{Basic Model: Enhanced Hierarchical Attention Bi-LSTM Network (HAN)}
Figure~\ref{fig:basic_model} illustrates the detailed architecture of our basic model comprising the two steps of sentence encoding and label prediction. Formally, a sentence encoding network returns sentence embeddings from pre-trained word embeddings. 
Then a label prediction network processes the sentence embeddings generated earlier 
and outputs the probabilities to indicate if sentences are the segment boundaries or not. Finally, to convert the numerical probabilities into binary labels, we follow the greedy decoding strategy in \citet{koshorek-etal-2018-text} by setting a threshold $\tau$. All the sentences with their probabilities over $\tau$ will be labeled $1$, and $0$ otherwise. 
This parameter $\tau$ is set in the validation stage.

For training, we compute the cross-entropy loss between the ground truth labels $Y = \{ y_1, 
... , y_{k-1} \}$ and our predicted probabilities $P = \{ p_1, 
... , p_{k-1} \}$ for a document with $k$ sentences:
\begin{equation}
    L_1 = -\sum_{i=1}^{k-1} [y_{i}\log p_{i} + (1-y_{i})\log (1-p_{i})] \label{eq:1}
\end{equation}
Looking at the details of the architecture in Figure~\ref{fig:basic_model}, our basic model constitutes a strong baseline by extending the segmenter presented in \citet{koshorek-etal-2018-text} in two ways (colored parts); namely, by improving the sentence encoder with an attention mechanism (orange) and with BERT embeddings (blue). 

\vspace{1ex}

\noindent
\textbf{Enhancing Task-Specific Sentence Representations -} While \citet{koshorek-etal-2018-text} applied max-pooling to build sentence embeddings from sentence encoding network, we applied an attention mechanism \cite{yang-etal-2016-hierarchical} to make the model better capture task-wise sentence semantics. The benefit of this enhancement is verified empirically by the results in Table~\ref{tab:bert}. As it can be seen, replacing the max-pooling with the attention based BiLSTM sentence encoder yields better performance.

\vspace{1ex}

\noindent
\textbf{Enhancing Generality with BERT Embeddings}
In order to better deal with unseen text in test data and hence improve the model's generality, we utilize a pre-trained BERT sentence encoder\footnote{ \url{%https://
github.com/hanxiao/bert-as-service}. For languages other than English, we use their corresponding pre-trained BERT models.} which complements 
our sentence encoding network. The transformer-based BERT model \cite{devlin-etal-2019-bert} was trained on multi-billion sentences publicly available on the web for several generic sentence-level semantic tasks, such as Natural Language Inference and Question Answering, 
which implies that it can arguably capture more general aspects of sentence semantics in a reliable way.
To combine task-specific information with generic semantic signals from BERT, we simply concatenate the BERT sentence embeddings with the sentence embeddings derived from our encoder. 
Such concatenation then becomes the input of the next level network (see Figure~\ref{fig:basic_model}). The benefit of injecting BERT embedding is also verified empirically by the results reported in Table~\ref{tab:bert}. We can see that concatenating BERT embedding and the output of Att-BiLSTM yields the best performance compared with only BERT embedding or the output of Att-BiLSTM.

\begin{table}
\scalebox{0.772}{
\begin{tabular}{l | c c c c}

\specialrule{.1em}{.05em}{.05em}

\textbf{Dataset} & \multicolumn{1}{c}{\textbf{CHOI}} & \multicolumn{1}{c}{\textbf{RULES}} & \multicolumn{1}{c}{\textbf{SECTION}} & \multicolumn{1}{c}{MEAN} \\
 \hline
 MaxPooling & 1.04 & 7.74 & 12.62 & 7.14 \\
 BiLSTM & 0.92 & 7.47 & 11.60 & 6.66 \\
 BERT & 0.93 & 8.35 & 12.08 & 7.12 \\
 BiLSTM+BERT & \textbf{0.81} & \textbf{6.90} & \textbf{11.30} & \textbf{6.34} \\
\specialrule{.1em}{.05em}{.05em}
\end{tabular}
}
\caption{\label{tab:bert} $P_k$ error score (lower the better, see Section~\ref{sec:solution:pk} for details) of different sentence encoding strategies on three datasets (Section~\ref{sec:solution:dataset}). To fit in the table, we shorten Att-BiLSTM to BiLSTM. Results in \textbf{bold} are the best performance across the comparisons.}
\end{table}

\begin{figure*}
\centering
\includegraphics[width=5.2in]{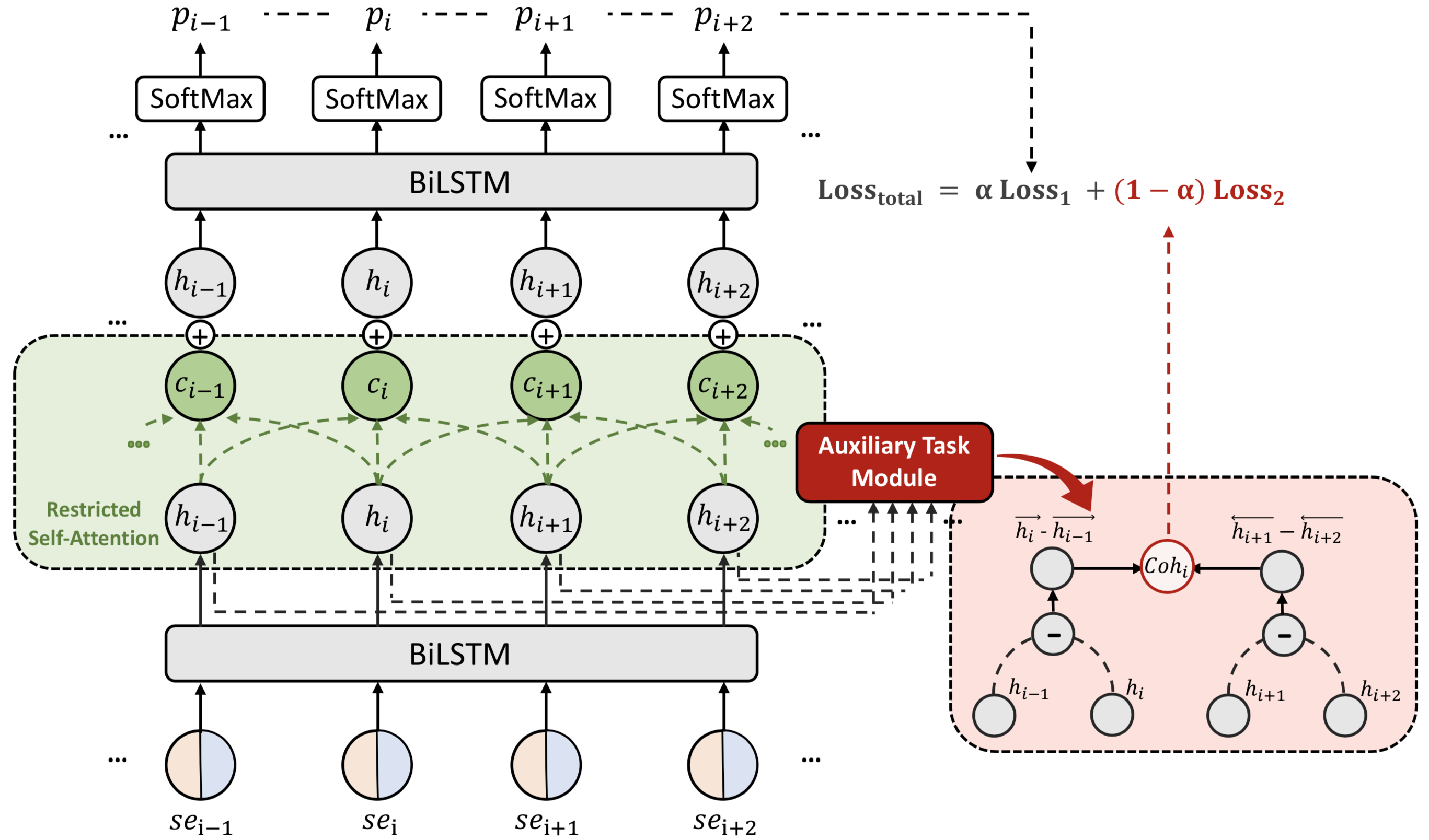}
\caption{\label{fig:fullmodel} 
Our full model with context modeling components: \textcolor{ao(english)}{restricted self-attention},  \textcolor{cardinal}{auxiliary task module}.}

\end{figure*}

\subsection{Auxiliary Task Learning}
In a well-structured document, 
the semantic coherence of a pair of sentences from the same segment should tend to be 
greater than the coherence of a pair of sentences from different segments. This observation provides us with an alternative way to enable better context modeling by formulating a coherence-related auxiliary task whose objective can be jointly optimized with our original objective (Equation~\ref{eq:1}).
This task thereby is to predict the consecutive sentence-pair coherence by using the sentence hidden states generated from the BiLSTM network. Concurrently minimizing the loss of this task can regulate our model to learn better semantic coherence relation between sentences by reducing the semantic coherence scores for the sentence pairs {\it across segments} and increasing the semantic coherence scores for the sentence pairs {\it within a segment}. 

To obtain the ground truth for our introduced auxiliary task (sentence-pair coherence prediction), we leverage the ground truth of our segmented training set rather than 
requiring external annotations. For a document which contains $m$ sentences, there are $m-1$ consecutive sentence pairs. If this document has $n$ segment boundaries, then among those $m-1$ sentence pairs, $n$ sentence pairs are from different segments, while the remaining $m-n-1$ sentence pairs are from the same segment. In order to concurrently minimize the coherence of the sentences from different segments and maximize the coherence of the sentences in the same segment, 
we give a sentence pair $<s_i, s_{i+1}>$ a coherence label $l_i=1$ if sentences in this pair are from the same segment, and $l_i=0$ otherwise.
The embeddings $e_{i}$ and $e_{i+1}$ of adjacent sentences pairs $<s_i , s_{i+1}>$ used for coherence computing are calculated from BiLSTM forward and backward hidden states $\overrightarrow{h}$ and $\overleftarrow{h}$, following the equations below: 
\begin{gather}
    e_i = \tanh(W_{e}(\overrightarrow{h_{i}} - \overrightarrow{h_{i-1}}) + b_{e})\\
    e_{i+1} = \tanh(W_{e}(\overleftarrow{h_{i+1}} - \overleftarrow{h_{i+2}}) + b_{e})
\end{gather}
However, notice that instead of using the conventional $[\overrightarrow{h_i}; \overleftarrow{h_i}]$ as the embedding of sentence $i$, here, similarly to \citet{wang-chang-2016-graph}, we subtract forward/backward states to focus on the semantics of sentences in the current sentence pair. 
The semantic coherence between two sentence embeddings is then computed as the sigmoid of their cosine similarity:
\begin{equation}
     Coh_i = \sigma(cos(e_i, e_{i+1}))
\end{equation}

We use binary cross-entropy loss to formulate the objective of our auxiliary task. For a document with $k$ sentences, the loss can be calculated as:
\begin{equation}
    L_2 = - \sum^{k-1}_{i=1, l_i=1}\log Coh_{i}-\sum^{k-1}_{i=1, l_i=0}\log(1-Coh_{i}) \label{eq:2}
\end{equation}
which penalizes high $Coh$ across segments and low $Coh$ within segments.

Combining Equation~\ref{eq:1} and~\ref{eq:2}, we form the loss function of our new segmenter as:
\begin{equation}
    L_{total} = \alpha L_{1} + (1 - \alpha) L_{2}
\end{equation}
with the trade-off parameter $\alpha$ tuned in validation stage, topic segmentation and the coherence-related auxiliary task are jointly optimized. The architecture of the auxiliary task module and its integration in our segmenter is shown in red in  Figure~\ref{fig:fullmodel}.

\begin{table*}
\centering
\scalebox{0.9}{
\begin{tabular}{c |@{\space\space\space\space\space} c@{\space\space\space\space\space} c@{\space\space\space\space\space} c@{\space\space\space\space\space} | c@{\space\space\space\space\space\space} c@{\space\space\space\space\space\space} c@{\space\space\space\space\space} c@{\space\space\space\space\space}}

\specialrule{.1em}{.05em}{.05em}

\textbf{Dataset} & \textbf{CHOI} & \textbf{RULES} & \textbf{SECTION} & \textbf{WIKI-50} & \textbf{CITIES} & \textbf{ELEMENTS} & \textbf{CLINICAL} \\
\hline
 documents & 920 & 4,461 & 21,376 & 50 & 100 & 118 & 227 \\
 \# sent/seg & 7.4 & 7.4 & 7.2 & 13.6 & 5.2 & 3.3 & 28.0\\
 \# seg/doc & 10.0 & 16.6 & 7.9 & 3.5 & 12.2 & 6.8 & 5.0\\
 real world & \includegraphics[width=0.02\textwidth]{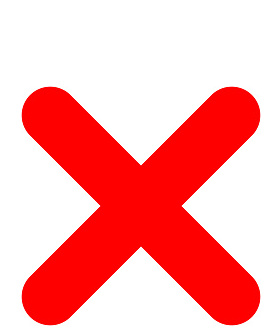} & \includegraphics[width=0.022\textwidth]{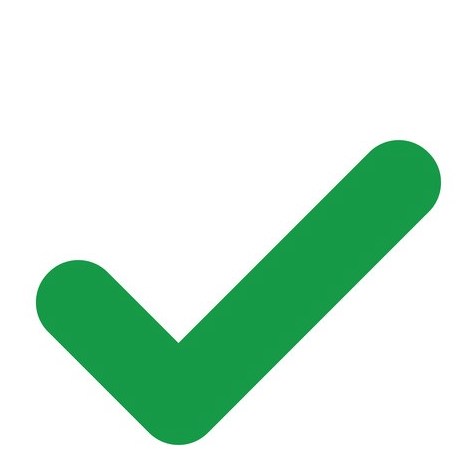} & \includegraphics[width=0.022\textwidth]{check.jpg} & \includegraphics[width=0.022\textwidth]{check.jpg} & \includegraphics[width=0.022\textwidth]{check.jpg} & \includegraphics[width=0.022\textwidth]{check.jpg} & \includegraphics[width=0.022\textwidth]{check.jpg} \\
\specialrule{.1em}{.05em}{.05em}
\end{tabular}
}
\caption{\label{tab:stats} Statistics of all the \textbf{English} topic segmentation datasets used in our experiments.}
\end{table*}

\begin{table}
\centering
\scalebox{0.93}{
\begin{tabular}{c | @{\space\space\space\space\space} c@{\space\space\space\space\space\space\space} c@{\space\space\space\space\space\space\space} c@{\space\space\space\space} }

\specialrule{.1em}{.05em}{.05em}

\textbf{Dataset} & \textbf{EN} & \textbf{DE} & \textbf{ZH}  \\
\hline
 documents & 21,376 & 12,993 & 10,000  \\
 \# sent/seg & 7.2 & 6.3 & 5.1 \\
 \# seg/doc & 7.9 & 7.0 & 6.4 \\
 real world & \includegraphics[width=0.02\textwidth]{check.jpg} & \includegraphics[width=0.022\textwidth]{check.jpg} & \includegraphics[width=0.022\textwidth]{check.jpg}  \\
\specialrule{.1em}{.05em}{.05em}
\end{tabular}
}
\caption{\label{tab:stats_multilingual} Statistics of the the WIKI-SECTION data in English(EN), German(DE) and Chinese(ZH).}
\end{table}

\subsection{Sentence-Level Restricted Self-Attention}

The self-attention mechanism \cite{NIPS2017_7181} has been widely applied to many sequence labeling tasks due to its superiority in modeling long-distance dependencies in text. However, when the task mainly requires modelling local context, long-distance dependencies will instead introduce noise. \citet{wang-etal-2018-toward} noticed this problem for discourse segmentation, where the crucial information for a clause-like Elementary Discourse Unit (EDU) boundary prediction comes usually only from the adjacent EDUs. Thus, they proposed a \textit{word-level} restricted self-attention mechanism by adding a fixed size window constraint on the standard self-attention. In essence, this mechanism encourages the model to absorb more information directly from adjacent context words within a fixed range of neighborhood. We hypothesize that the
similar restricted dependencies also play a dominant role in 
topic segmentation due to their close relation. Hence, instead of at word-level, 
we add a \textit{sentence-level} restricted self-attention on top of the label prediction network of the basic model, as shown in green in Figure~\ref{fig:fullmodel}.

In particular, once hidden states are obtained for all the sentences of document $d$, we 
compute the similarities between the current sentence $i$ and its nearby sentences within a window of size $S$. For example, the similarity between sentence $s_i$ and $s_{j}$ which is within the window size is computed as:
\begin{equation}
    sim_{i,j} = W_{a}[h_{i};h_{j};(h_{i} \odot h_{j})] + b_a
\end{equation}
where $h_i$, $h_j$ are the hidden state of $s_i$ and $s_j$. $W_a$ and $b_a$ are both attention parameters. $;$ is the concatenation operation and $\odot$ is the dot product operation. The attention weights for all the sentences in the fixed window are:
\begin{equation}
    a_{i,j} = \frac{e^{sim_{i,j}}}{\sum_{s=-S}^{S} e^{sim_{i,i+s}}}
\end{equation}
The output for sentence $i$ after the restricted self-attention mechanism is the weighted sum of all the sentence hidden states within the window:
\begin{equation}
    c_{i} = \sum_{s=-S}^{S} a_{i,i+s}h_{i+s}
\end{equation}
where $c_i$ denotes the \textit{local context embedding} of sentence $i$ generated by restricted self-attention. After getting the local context embeddings for all the sentences, we concatenate them with the original sentence hidden states and input them to another BiLSTM layer (top of Figure \ref{fig:fullmodel}).

\section{Experimental Setup}
In order to comprehensively evaluate the effectiveness of our context modeling strategy of adding a coherence-related auxiliary task and a restricted self-attention mechanisms to the basic model, we conduct three sets of experiments for evaluation: 
{\small \textbf{\texttt{(i) Intra Domain 
}}}: we train and test the models in the same domain, repeating this evaluation for three different domains (datasets).
{\small \textbf{\texttt{(ii) Domain Transfer 
}}}: we train the models on a large dataset which covers a variety of topics and test them on four challenging real-world datasets.  
{\small \textbf{\texttt{(iii) Multilingual 
}}}: we train and test our model on three datasets within different languages (English, German and Chinese), to assess our proposed strategy’s generality within different languages.
\subsection{Datasets}
\label{sec:solution:dataset}
\textbf{Data for Intra-Domain Evaluation} 
High quality training dataset for topic segmentation usually satisfies the following criteria: (1) large size; (2) cover a variety of topics; (3) contains real documents with reliable segmentation either from human annotations or already specified in the documents e.g., sections. In order to comprehensively evaluate the effectiveness of our context modeling strategy when dealing with 
data of different quality, we train and test models on the following three datasets: \\
\textbf{\textit{CHOI}} \cite{choi-2000-advances} whose articles are synthesized artificially by stitching together different sources (i.e., they were not written as one document by one author). Hence, it does not really 
reflect naturally occurring topic drifts. While the quality of this dataset is low, it is an early but popular benchmark for topic segmentation evaluation. We include this dataset to allow comparison with the previous work. \\
\textbf{\textit{RULES}} \cite{bertrand2018hall} is a dataset collected from the U.S. Federal Register issues\footnote{\url{www.govinfo.gov/}}.
When U.S. federal agencies make changes to regulations or other policies, they must publish a document called a “Rule” in the Federal Register. The Rule describes what is being changed and discusses the motivation and legal justification for the action. Since each paragraph in a document discusses one topic, we consider the last sentence of each paragraph as a ground truth topic boundary.
The discussion paragraphs usually cover diverse topics in formal, technical language that can be hard to find online, so we deem it as an additional well-labelled dataset for testing topic segmentation to complement our other datasets which contain more informal use of the language. \\
\textbf{\textit{WIKI-SECTION}} \cite{arnold2019sector} is a newly released dataset which was originally generated from the most recent English and German Wikipedia dumps. To better align with the purpose of intra-domain experiment, we only select the English samples for training and the German samples will be used in the experiments of multilingual evaluation. The English \textit{WIKI-SECTION} (labeled \textit{SECTION} in the tables) consists of Wikipedia articles from domain \textit{diseases} and \textit{cities}. 
We deem this dataset as the most reliable training source among the three datasets. It has the largest size and the two domains (\textit{cities} and \textit{diseases}) cover news-based samples and scientific-based samples respectively. 

We split \textit{CHOI} and \textit{RULES} into 80\% training, 10\% validation, 10\% testing. For \textit{SECTION}, we follow \citet{arnold2019sector} and split it into 70\% training, 10\% validation, 20\% testing. Table~\ref{tab:stats} (left) contains the statistical details for these three sets.

\vspace{1ex}

\noindent
\textbf{Data for Domain Transfer Evaluation} We pick \textit{WIKI-SECTION} as our training set in this line of experiments, due to its largest size and variety of covered topics. Following previous work, we evaluate our model and baselines on four datasets that originate from different source distributions:
{\textbf{\textit{WIKI-50}} \cite{koshorek-etal-2018-text}} which consists of 50 samples randomly generated from the latest English Wikipedia dump, with no overlap with training and validation data.
{\textbf{\textit{Cities}} \cite{chen-etal-2009-global}} which consists of 100 samples generated from Wikipedia about cities. We also ensure that this dataset has no overlap with training and validation data.
{\textbf{\textit{Elements}} \cite{chen-etal-2009-global}} which consists of 118 samples generated from Wikipedia about chemical elements.
{\textbf{\textit{Clinical Books}} \cite{malioutov-barzilay-2006-minimum}} which consists of 227 chapters from a medical textbook.
Table~\ref{tab:stats} (right) gives more detailed statistics for these datasets.

\vspace{1ex}

\noindent
\textbf{Data For Multilingual Evaluation} In order to test the effectiveness of our 
context modeling strategy across languages, 
besides the English \textit{WIKI-SECTION}, we train and test our model on two other Wikipedia datasets in German and Chinese: 
\\
\textbf{\textit{SECTION-DE}} which was released together with English \textit{WIKI-SECTION} in \citet{arnold2019sector}. It also contains articles about cities and diseases. The section marks are used as the ground truth labels. \\ 
\textbf{\textit{SECTION-ZH}} which was randomly generated from the Chinese Wikipedia dump\footnote{\url{https://linguatools.org/tools/corpora/wikipedia-monolingual-corpora/}} mentioned in \citet{wiki_source}. 
As before, section marks are also used here as ground truth boundaries. The statistical details of these two datasets can be found in Table~\ref{tab:stats_multilingual}.

\subsection{Baselines}
These include two popular unsupervised topic segmentation methods, \textit{BayesSeg} \cite{eisenstein-barzilay-2008-bayesian} and \textit{GraphSeg} \cite{glavas-etal-2016-unsupervised}, as well as the three recently proposed 
supervised neural models, \textit{TextSeg} \cite{koshorek-etal-2018-text} (from which we derive our basic model), \textit{Sector} \cite{arnold2019sector} and \textit{Hierarchical Transformer} (labeled \textit{Transformer} in the tables) \cite{glavas-2020-two}. We use the original implementation of \textit{BayesSeg}, \textit{GraphSeg} and \textit{TextSeg}. We reimplement the \textit{Hierarchical Transformer} ourselves. In Table~\ref{tab:res_transfer}, we adopt the results of \textit{BayesSeg}, \textit{GraphSeg} and \textit{Sector} from \citet{arnold2019sector}\footnote{\citet{arnold2019sector} reported \textit{Sector}'s performance on multiple model settings. Here we pick the performance of the model trained on wikifull to be close to our training setting.}.

\begin{table}
\centering
\scalebox{0.81}{
\begin{tabular}{l | c c c c}

\specialrule{.08em}{.05em}{.05em}

\textbf{Dataset} & \multicolumn{1}{c}{\textbf{CHOI}} & \multicolumn{1}{c}{\textbf{RULES}} & \multicolumn{1}{c}{\textbf{SECTION}} & \multicolumn{1}{c}{MEAN} \\
 \hline
 Random & 49.4\textsuperscript{\space} & 50.6\textsuperscript{\space} & 51.3\textsuperscript{\space} & 50.4\textsuperscript{\space} \\
\hline
 BayesSeg & 20.8\textsuperscript{\space} & 41.5\textsuperscript{\space} & 39.5\textsuperscript{\space} & 33.9\textsuperscript{\space} \\
 GraphSeg & 6.6\textsuperscript{\space} & 39.3\textsuperscript{\space} & 44.9\textsuperscript{\space} & 30.3\textsuperscript{\space} \\
 TextSeg & 1.0\textsuperscript{\space} & 7.7\textsuperscript{\space} & 12.6\textsuperscript{\space} & 7.1\textsuperscript{\space} \\
 Sector  & -\textsuperscript{\space} & -\textsuperscript{\space} & 12.7\textsuperscript{\space} & -\textsuperscript{\space}\\
 Transformer & 4.8\textsuperscript{\space} & 9.6\textsuperscript{\space} & 13.6\textsuperscript{\space} & 9.3\textsuperscript{\space}\\
 \hline
 Basic Model& 0.81\textsuperscript{\space} & 7.0\textsuperscript{\space} & 11.3\textsuperscript{\space} & 6.4\textsuperscript{\space} \\
 +AUX & 0.64\textsuperscript{\textdagger} & 6.1\textsuperscript{\textdagger} & 10.4\textsuperscript{\textdagger} & 5.7\textsuperscript{\space}\\
 +RSA & 0.72\textsuperscript{\textdagger} & 6.3\textsuperscript{\textdagger} & 10.0\textsuperscript{\textdagger} & 5.7\textsuperscript{\space}\\
 +AUX+RSA & \underline{\textbf{0.54}}\textsuperscript{\textdagger} & \underline{\textbf{5.8}}\textsuperscript{\textdagger} & \underline{\textbf{9.7}}\textsuperscript{\textdagger} & \underline{\textbf{5.3}}\textsuperscript{\space}\\
\specialrule{.08em}{.05em}{.05em}
\end{tabular}
}
\caption{\label{tab:res_intra_domain} $P_k$ error score on three datasets. Results in \textbf{bold} indicate the best performance across all comparisons. \underline{Underlined} results indicate the best performance in the bottom section. $\dagger$ indicates the result is significantly different ($p<0.05$) from basic model. }
\end{table}

\subsection{Evaluation Metric}
\label{sec:solution:pk}
 We use the standard $P_k$ error score \cite{Beeferman1999} as our evaluation metric,  since it has become the standard for comparing topic segmenters.
$P_k$ is calculated as:
\begin{align*}
\resizebox{1\hsize}{!}{
    $P_{k}(ref,hyp) = \sum_{i=0}^{n-k}\delta_{ref}(i,i+k) \neq \delta_{hyp}(i,i+k)$
}
\end{align*}
where $\delta$ is an indicator function which is 1 if sentence $i$ and $i+k$ are in the same segment, 0 otherwise.
It measures the probability of mismatches between the ground truth segments (\textit{ref}) and model predictions (\textit{hyp}) within a sliding window $k$. As a standard setting which has been used in previous work, window size $k$ is the average segment length of \textit{ref}.
Since $P_k$ is a penalty metric, lower score indicates better performance.

\subsection{Neural Model Setup}
Following \citet{koshorek-etal-2018-text}, our initial word embeddings are GoogleNews word2vec ($d = 300$). We also use word2vec embeddings ($d = 300$) and Fasttext embeddings ($d = 300$), which are both derived from Wikipedia corpora for German and Chinese respectively. We use the Adam optimizer, setting the learning rate to 0.001  and batch size to 8. The BiLSTM hidden state size is 256 following 
\citet{koshorek-etal-2018-text}. Model training is done for 10 epochs and performance is monitored over the validation set. We generate BERT sentence embeddings with the pre-trained 12-layer model released by Google AI (embedding size 768). The window size of restricted self-attention is 3 and $\alpha$ is 0.8. These were tuned on the validation sets of the datasets we use.

\begin{table}
\centering
\scalebox{0.819}{
\begin{tabular}{l | c c c c}

\specialrule{.1em}{.05em}{.05em}

\textbf{Dataset} & \multicolumn{1}{c}{\textbf{Wiki-50}} & \multicolumn{1}{c}{\textbf{Cities}} & \multicolumn{1}{c}{\textbf{Elements}} & \multicolumn{1}{c}{\textbf{Clinical}} \\
 \hline
 Random & 52.7\textsuperscript{\space} & 47.1\textsuperscript{\space} & 50.1\textsuperscript{\space} & 44.1\textsuperscript{\space} \\
\hline
 BayesSeg & 49.2\textsuperscript{\space} & 36.2\textsuperscript{\space} & \textbf{35.6}\textsuperscript{\space} & 57.2\textsuperscript{\space} \\
 GraphSeg & 63.6\textsuperscript{\space} & 40.0\textsuperscript{\space} & 49.1\textsuperscript{\space} & 64.6\textsuperscript{\space} \\
 TextSeg & 28.5\textsuperscript{\space} & 19.8\textsuperscript{\space} & 43.9\textsuperscript{\space} & 36.6\textsuperscript{\space} \\
 Sector  & 28.6\textsuperscript{\space} & 33.4\textsuperscript{\space} & 42.8\textsuperscript{\space} & 36.9\textsuperscript{\space}\\
 Transformer & 29.3\textsuperscript{\space} & 20.2\textsuperscript{\space} & 45.2\textsuperscript{\space} & 35.6\textsuperscript{\space}\\
 \hline
 Basic Model& 28.7\textsuperscript{\space} & 17.9\textsuperscript{\space} & 43.5\textsuperscript{\space} & 33.8\textsuperscript{\space} \\
 +AUX & 27.9\textsuperscript{\space} & 17.0\textsuperscript{\textdagger} & 41.8\textsuperscript{\textdagger} & 31.5\textsuperscript{\textdagger}\\
 +RSA & 27.8\textsuperscript{\textdagger} & 16.8\textsuperscript{\textdagger} & 42.7\textsuperscript{\space} & 31.9\textsuperscript{\textdagger}\\
 +AUX+RSA & \underline{\textbf{26.8\textsuperscript{\textdagger}}} & \underline{\textbf{16.1\textsuperscript{\textdagger}}} & \underline{39.4\textsuperscript{\textdagger}} & \underline{\textbf{30.5\textsuperscript{\textdagger}}}\\
\specialrule{.1em}{.05em}{.05em}
\end{tabular}
}
\caption{\label{tab:res_transfer} $P_k$ error score on four test sets. Results in \textbf{bold} indicate the best performance across all comparisons. \underline{Underlined} results indicate the best performance in the bottom section. $\dagger$ indicates the result is significantly different ($p<0.05$) from basic model.}
\end{table}

\section{Results and Discussion}
\subsection{Intra-Domain Evaluation}
Table~\ref{tab:res_intra_domain} shows the 
models' performance on the three datasets, 
when all supervised models are trained and evaluated on the training and test set from the same domain. 
To investigate the effectiveness of auxiliary task (AUX) and restricted self-attention (RSA), Table~\ref{tab:res_intra_domain} also shows the results of individually adding each component to our basic segmenter.
The most important observation from the table is that our model enhanced by context modeling outperforms all the supervised and unsupervised baselines with a substantial performance gain.
With our context modeling strategy, the average $P_k$ scores of our model over the three datasets improves on the best model (TextSeg) among the baselines by 25\%. Compared with the basic model, adding AUX or RSA equally gives significant and consistent improvement across all three sets. Adding both AUX and RSA results in the biggest improvement by up to 17\% on the mean across the three datasets.

\subsection{Domain Transfer Evaluation}
Table~\ref{tab:res_transfer} compares the performance of the baselines and our model on four challenging real-world test datasets. All supervised models are trained on the training set of \textit{WIKI-SECTION}.
One important observation 
is that our model enhanced by context modeling outperforms all the baseline methods on three out of four test sets with a substantial performance gap. Admittedly, \textit{BayesSeg} performs better on \textit{Elements}, 
possibly because that merely word embedding similarity is sufficient to indicate 
segment boundaries in this dataset. However, \textit{BayesSeg} 
is completely dominated by our model on the other test sets. Overall, this 
indicates that our proposed context modeling strategy can not only enhance the model under the intra-domain setting, but also produce robust models that transfer to other unseen domains.
Furthermore, we observe that  AUX 
 and RSA 
are both necessary for our model, since they do not only improve performance individually, but they achieve the best results when synergistically combined. 

\begin{table}
\centering
\scalebox{0.95}{
\begin{tabular}{l@{\space\space\space} | @{\space\space\space\space} c@{\space\space\space\space\space\space\space\space} c@{\space\space\space\space\space\space\space\space} c@{\space\space\space\space}}

\specialrule{.1em}{.05em}{.05em}
\textbf{Dataset} & \textbf{EN} & \textbf{DE} & \textbf{ZH}  \\
 \hline
 Random & 51.3\textsuperscript{\space} & 48.7\textsuperscript{\space} & 52.2\textsuperscript{\space} \\
\hline
 Basic Model& 11.3\textsuperscript{\space} & 18.2\textsuperscript{\space} & 20.5\textsuperscript{\space} \\
 +AUX & 10.4\textsuperscript{\textdagger} & 17.7\textsuperscript{\space} & 20.5\textsuperscript{\space}\\
 +RSA & 10.0\textsuperscript{\textdagger} & 16.6\textsuperscript{\textdagger} & \textbf{19.8}\textsuperscript{\textdagger} \\
 +AUX+RSA & \textbf{9.7}\textsuperscript{\textdagger} & \textbf{15.9}\textsuperscript{\textdagger} & 20.0\textsuperscript{\textdagger} \\
\specialrule{.1em}{.05em}{.05em}
\end{tabular}
}
\caption{\label{tab:res_multilingual} $P_k$ error score on the datasets in three languages (English, German and Chinese). }
\end{table}

\subsection{Multilingual Evaluation}
Table~\ref{tab:res_multilingual} shows results for our context modeling strategy across three different languages: English (EN), German (DE) and Chinese (ZH). Remarkably, even our basic model without any add-on component outperforms the random baseline by a wide margin. Looking at the gains from AUX and RSA, for German we observe a pattern similar to English, with our complete context modeling strategy (AUX+RSA) delivering the strongest gains. However, the performance on Chinese is not as strong as on English and German. Employing RSA still achieves a statistically significant 0.7 $P_k$ score drop, but introducing AUX does not help. One possible reason is that the sentences in the Chinese Wikipedia pages are relatively short and fragmented. Thus, the semantics of these sentences may be too simple to sufficiently guide the coherence auxiliary task. In general, when comparing the behavior of our context modeling strategy across these three languages, RSA appears to yield stable benefits, while the effectiveness of AUX seems to depend more on peculiarities of the dataset in the target language.

\section{Conclusions and Future Work}
We address a serious limitation of current neural topic segmenters, namely their inability to effectively model context. To this end, we propose a novel neural model that adds a coherence-related auxiliary task and restricted self-attention on top of a hierarchical BiLSTM attention segmenter 
to make better use of the contextual information. Experimental results of intra-domain on three datasets show that our strategy is effective within domains. Further, results on four challenging real-world benchmarks demonstrate its effectiveness in domain transfer settings. Finally, the application to other two languages (German and Chinese) suggests that our strategy has its potential in multilingual scenarios. 

As future work, we will investigate whether our proposed context modeling strategy is also effective for segmenting dialogues \cite{ijcai2018} rather than just standard articles. 
Secondly, we will explore how to capture even more accurate and informative contextual information by integrating document structures or sentence dependencies obtained from other NLP tasks (e.g., discourse parsing \cite{huber-carenini-2019-predicting, patrick2020} or discourse role labeling \cite{zeng-etal-2019-say}).

\section*{Acknowledgments}
\vspace{-1mm}
We thank the anonymous 
reviewers and the UBC-NLP group for their insightful comments.

\bibliography{aacl-ijcnlp2020}
\bibliographystyle{acl_natbib}

\end{document}